%% file: main.tex
\documentclass[10pt,twocolumn,letterpaper]{article}

\usepackage{iccv}              


\definecolor{iccvblue}{rgb}{0.21,0.49,0.74}
\usepackage[pagebackref,breaklinks,colorlinks,allcolors=iccvblue]{hyperref}

\usepackage{algorithm}
\usepackage{multirow}
\usepackage{multicol}

\usepackage{booktabs}
\usepackage{adjustbox}
\usepackage{algpseudocode}
\def\paperID{*****} 
\def\confName{ICCV}
\def\confYear{2025}
\usepackage{float}

\title{DAC-LoRA: Dynamic Adversarial Curriculum for Efficient and Robust Few-Shot Adaptation}

\author{
Ved Umrajkar\\
Indian Institute of Technology, Roorkee\\
{\tt\small v\_umrajkar@ma.iitr.ac.in}
}

\begin{document}
\maketitle
\input{sec/0_abstract}    
\input{sec/1_intro}
\input{sec/prelim}

\input{sec/method}

\input{sec/results}
{
    \small
    \bibliographystyle{ieeenat_fullname}
    \bibliography{main}
}

\input{sec/X_suppl}

\end{document}

%% file: sec/0_abstract.tex
\begin{abstract}
Vision-Language Models (VLMs) are foundational to critical applications like autonomous driving, medical diagnosis, and content moderation. While Parameter-Efficient Fine-Tuning (PEFT) methods like LoRA enable their efficient adaptation to specialized tasks, these models remain vulnerable to adversarial attacks that can compromise safety-critical decisions. CLIP, the backbone for numerous downstream VLMs, is a high-value target whose vulnerabilities can cascade across the multimodal AI ecosystem.We propose Dynamic Adversarial Curriculum \textbf{(DAC-LoRA)}, a novel framework that integrates adversarial training into PEFT. The core principle of our method---an intelligent curriculum of progressively challenging attacks---is general and can potentially be applied to any iterative attack method. Guided by the First-Order Stationary Condition (FOSC) and a TRADES-inspired loss, DAC-LoRA achieves substantial improvements in adversarial robustness without significantly compromising clean accuracy. Our work presents an effective, lightweight, and broadly applicable method to demonstrate that the DAC-LoRA framework can be easily integrated into a standard PEFT
pipeline to significantly enhance robustness.

\end{abstract}

%% file: sec/1_intro.tex
\section{Introduction}

Vision-Language Models (VLMs) have revolutionized computer vision by learning rich, transferable representations from vast amounts of image-text data~\cite{radford2021clip, jia2021scaling, li2023blip}. Their ability to perform zero-shot classification on unseen tasks has made them a cornerstone of modern AI~\cite{zanella2024lora-vlm}. However, to achieve state-of-the-art performance on specialized, downstream tasks, these powerful models require adaptation.

Full fine-tuning of such massive models is often computationally prohibitive. This has spurred the development of Parameter-Efficient Fine-Tuning (PEFT) techniques ~\cite{peft1,peft2,peft3,peft4, peft5, hu2021lora}, some of which adapt a model by training only a small fraction of its parameters~\cite{hu2021lora, peft3, peft7}. Among these, Low-Rank Adaptation (LoRA) has emerged as a particularly effective and efficient method~\cite{hu2021lora}. Recent work shows that applying LoRA to CLIP (CLIP-LoRA) is a powerful strategy for few-shot adaptation, achieving impressive accuracy on new tasks with minimal training and no task-specific hyperparameter tuning~\cite{zanella2024lora-vlm}.

Despite their accuracy on clean data, a critical and often overlooked aspect of fine-tuned VLMs is their vulnerability to \textbf{adversarial attacks}~\cite{goodfellow2015fgsm, madry2018pgd}. These attacks use small, human-imperceptible perturbations to cause the model to elicit incorrect classifications with high confidence. Crucially, standard fine-tuning on clean examples does not inherently confer robustness~\cite{ji2025enhancingadversarialrobustnessvisionlanguage}, leaving these powerful models fragile and unreliable in security-sensitive applications.

To address this gap, our work integrates robust adversarial training directly into the parameter-efficient LoRA framework. Instead of fine-tuning only on clean data or with fixed-strength attacks, our core contribution is the use of a \textbf{dynamic curriculum of adversarial examples}. Inspired by findings that a gradually increasing attack difficulty is more effective~\cite{wang2019fosc}, we use the \textbf{First-Order Stationary Condition (FOSC)} to intelligently control the attack potency throughout training. This allows the model to first build a foundation of robustness against simpler attacks before being exposed to more challenging ones. The result is DAC-LoRA, a method that produces an accurate, efficient, and robustly adapted model.

%% file: sec/prelim.tex
\begin{figure*}[t]
    \centering
    \includegraphics[width=0.9\linewidth]{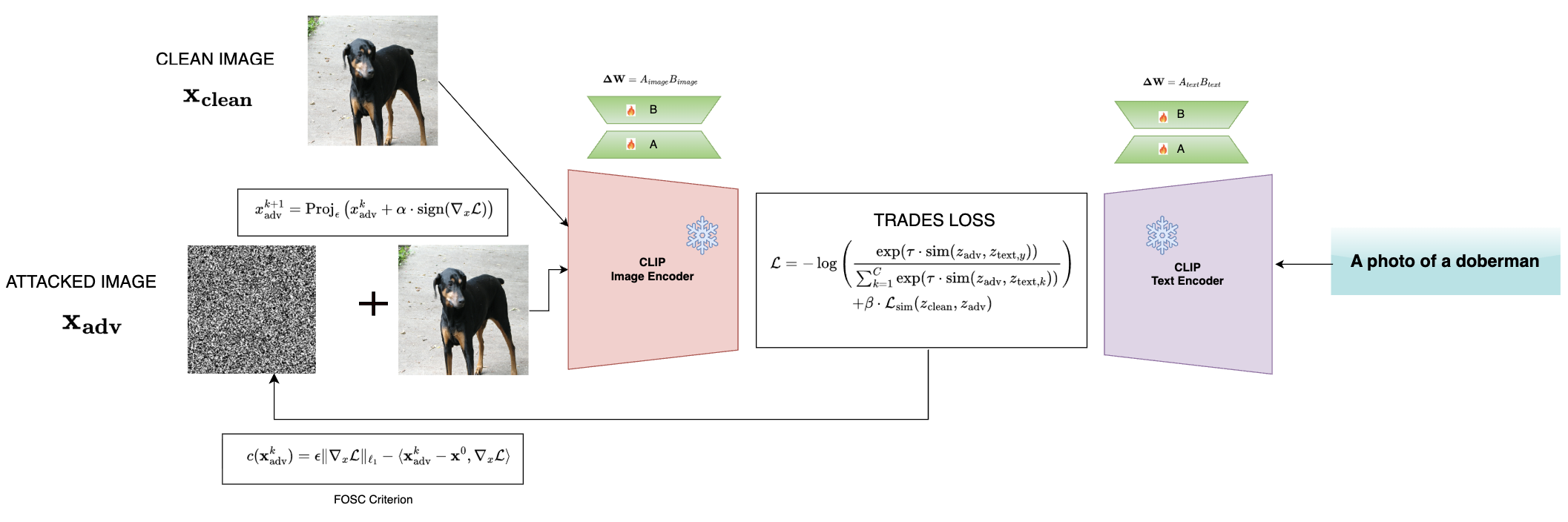} 
    \caption{Adversarial examples ($\mathbf{x}_{\text{adv}}$) are generated from clean images ($\mathbf{x}_{\text{clean}}$) using a PGD attack, with its potency dynamically controlled by the FOSC criterion. A TRADES-inspired loss, computed from the LoRA-adapted image and text encoders, is used to update only the low-rank matrices ($\mathbf{A}, \mathbf{B}$), while the original CLIP weights remain frozen.}
    \label{fig:workflow}
\end{figure*}

\section{Preliminaries}
\label{sec:preliminaries}

Our approach integrates parameter-efficient fine-tuning with a novel adversarial training strategy. This section reviews the four foundational concepts upon which our work is built: Low-Rank Adaptation (LoRA), standard adversarial attack generation, the TRADES framework for robust optimization, and the FOSC criterion for measuring attack potency.

\subsection{Few-Shot Adaptation with LoRA}
To adapt large pre-trained models without incurring the prohibitive cost of full fine-tuning, we use the Parameter-Efficient Fine-Tuning (PEFT) technique known as Low-Rank Adaptation (LoRA)~\cite{hu2021lora}. LoRA injects small, trainable low-rank matrices into the existing layers of a model. For a given weight matrix $\mathbf{W}$, its update is approximated by two smaller matrices, $\mathbf{A}$ and $\mathbf{B}$, such that the forward pass becomes $h = \mathbf{W}\mathbf{x} + \gamma \mathbf{B}\mathbf{A}\mathbf{x}$. This is highly effective for the few-shot adaptation of CLIP~\cite{radford2021clip,zanella2024lora-vlm}, as only the LoRA matrices ($\mathbf{A}, \mathbf{B}$) are trained, leaving the vast majority of the original model parameters frozen.

\subsection{Adversarial Attack Methods}
Adversarial attacks seek to generate a small, imperceptible perturbation $\boldsymbol{\delta}$ that, when added to a clean image $\mathbf{x}^0$, maximizes the model's loss. We use the following standard attack method to generate adversarial examples for our training curriculum.

\textbf{Projected Gradient Descent (PGD)} is a powerful, iterative attack widely used as a benchmark for evaluating model robustness~\cite{madry2018pgd}. It takes multiple small steps of size $\alpha$ in the direction of the gradient sign. After each step, it projects the result back into the $\epsilon$-ball around the original image to ensure the perturbation remains within the allowed magnitude:
\begin{equation}
    \mathbf{x}^{k+1} = \text{Proj}_{\|\mathbf{x} - \mathbf{x}^0\|_{\infty} \le \epsilon} \left( \mathbf{x}^k + \alpha \cdot \text{sign}(\nabla_{\mathbf{x}} \mathcal{L}(h_{\boldsymbol{\theta}}(\mathbf{x}^k), y)) \right)
    \label{eq:pgd}
\end{equation}

\subsection{Adversarial Robustness with TRADES}
The goal of adversarial training is to solve a min-max optimization problem that balances performance on clean data with robustness against attacks. The \textbf{TRADES} framework provides a principled method for this by decomposing the objective into a clean loss and a robustness-regularization term~\cite{zhang2019trades}. The original TRADES loss is:
\begin{equation}
    \mathcal{L}_{\text{TRADES}} = \mathcal{L}(h(\mathbf{x}^0), y) + \beta \cdot \text{KL}(h(\mathbf{x}^0) || h(\mathbf{x}_{\text{adv}}))
    \label{eq:trades}
\end{equation}
where KL is the Kullback-Leibler divergence. Our work uses a \textbf{TRADES-inspired variant} where the KL-divergence term is replaced with a cosine similarity loss between the feature embeddings of the clean and adversarial images. This serves a similar purpose by regularizing the model's feature space, encouraging it to produce consistent representations even when the input is perturbed.

\subsection{First-Order Stationary Condition (FOSC)}
\label{sec:fosc_prelim}

To implement a \textit{dynamic} curriculum, we need a way to measure the strength of an adversarial example. The \textbf{First-Order Stationary Condition (FOSC)} provides a principled criterion to quantify how well the inner maximization problem of an attack has been solved~\cite{wang2019fosc}. For an adversarial example $\mathbf{x}^k$, the FOSC score quantifies the potential to further increase the model's loss. The closed-form solution is:
\begin{equation}
    c(\mathbf{x}^k) = \epsilon \|\nabla_{\mathbf{x}} \mathcal{L}(h(\mathbf{x}^k), y)\|_{1} - \langle \mathbf{x}^k - \mathbf{x}^0, \nabla_{\mathbf{x}} \mathcal{L}(h(\mathbf{x}^k), y) \rangle
    \label{eq:fosc}
\end{equation}
We provide an intuitive derivation and further details on FOSC in the Supplementary Material (see Section~\ref{sec:supp_fosc}). A \textbf{lower FOSC score indicates a stronger}, better-converged adversarial example~\cite{wang2019fosc}. This criterion is the cornerstone of our dynamic curriculum, enabling us to precisely control attack difficulty during training.

%% file: sec/method.tex
\begin{figure*}
    \centering
    \includegraphics[width=0.9\linewidth]{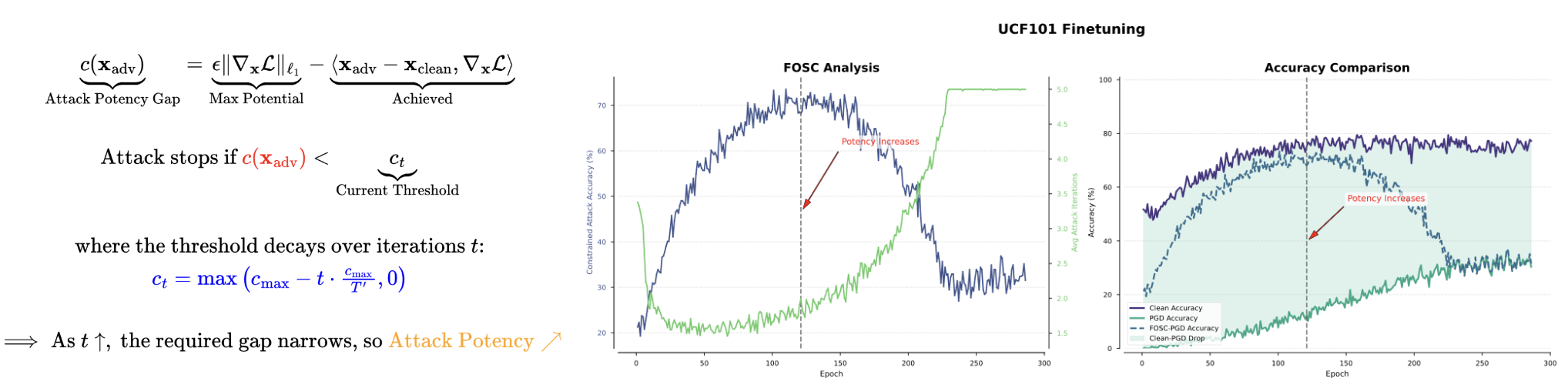}
    \caption{Training dynamics of our DAC-LoRA framework on the UCF101 dataset, illustrating the principles of our dynamic curriculum. Unlike a fixed-strength PGD attack, our FOSC-PGD method dynamically adjusts attack potency based on a decaying threshold, as detailed mathematically on the left. The middle plot visualizes this curriculum, showing how the required attack strength (green line) increases during training. The result, shown on the right, is a significant improvement in adversarial robustness (light blue line).}
    \label{fig:placeholder}
\end{figure*}

\section{Methodology: Adversarial LoRA with a Dynamic Curriculum}
\label{sec:methodology}

Our method synergizes the efficiency of LoRA with a robust training objective by fine-tuning the LoRA parameters on a dynamic curriculum of adversarial examples. The overall workflow is illustrated in Figure~\ref{fig:workflow}.

\subsection{Robust Fine-Tuning Objective}
We adapt the standard adversarial training objective for the PEFT setting. Our goal is to find the optimal LoRA parameters $\boldsymbol{\theta}_{\text{LoRA}}$ that minimize the loss on adversarially perturbed data, while the pre-trained CLIP parameters $\boldsymbol{\theta}_{\text{CLIP}}$ remain frozen. We employ a TRADES-inspired loss function that balances performance on adversarial examples with feature consistency:
\begin{equation}
\min_{\boldsymbol{\theta}_{\text{LoRA}}} \mathbb{E}_{(\mathbf{x}^0, y) \sim \mathcal{D}} \left[ \mathcal{L}(h(\mathbf{x}_{\text{adv}}), y) + \beta \cdot \mathcal{L}_{\text{sim}}(h(\mathbf{x}^0), h(\mathbf{x}_{\text{adv}})) \right]
\label{eq:robust_objective}
\end{equation}
where $\mathbf{x}_{\text{adv}}$ is the generated adversarial example, $\mathcal{L}$ is the cross-entropy loss, and $\mathcal{L}_{\text{sim}}$ is a cosine similarity loss between the feature embeddings of the clean ($\mathbf{x}^0$) and adversarial images.

\subsection{The Dynamic FOSC Attack Curriculum}
Instead of using a fixed-strength PGD attack, we implement a dynamic curriculum based on the FOSC criterion (see Section~\ref{sec:fosc_prelim}). This allows us to start with weaker attacks and gradually increase their potency as the model becomes more robust. We define a dynamic FOSC threshold $c_t$ that decays linearly over training iterations $t$:
\begin{equation}
    c_t = \max(c_{\max} - t \cdot \frac{c_{\max}}{T'}, 0)
    \label{eq:fosc_decay}
\end{equation}
where $c_{\max}$ is a high initial value and $T'$ is the curriculum duration. The PGD attack on an example is halted as soon as its FOSC score (from Eq.~\ref{eq:fosc}) falls below this threshold. This ensures the model learns robust features in a stable and progressive manner.

\subsection{The DAC-LoRA Algorithm}
Our complete procedure is outlined in Algorithm~\ref{alg:dac_lora}. At each training step, we first generate a batch of adversarial examples using our FOSC-controlled PGD attack. Then, we compute our robust loss function and use the resulting gradients to update \textit{only} the LoRA parameters.

\begin{algorithm}[t]
\caption{DAC-LoRA Fine-Tuning}
\label{alg:dac_lora}
\begin{algorithmic}[1]
\State \textbf{Input:} Pre-trained model $h_{\boldsymbol{\theta}_{\text{CLIP}}}$, data $S$, LoRA params $\boldsymbol{\theta}_{\text{LoRA}}$.
\State \textbf{Hyperparameters:} Iterations $T$, LR $\eta_t$, FOSC params $c_{\max}, T'$.
\For{$t = 0$ to $T-1$}
    \State $c_t \leftarrow \max(c_{\max} - t \cdot c_{\max} / T', 0)$ 
    \For{each batch $(\mathbf{x}_{\mathcal{B}}^0, y_{\mathcal{B}})$ from $S$}
        \State $\mathbf{x}_{\text{adv}} \leftarrow \text{FOSC-PGD}(h, \mathbf{x}_{\mathcal{B}}^0, y_{\mathcal{B}}, c_t)$ 
        \State $\mathcal{L}_{\text{batch}} \leftarrow \mathcal{L}(h(\mathbf{x}_{\text{adv}}), y_{\mathcal{B}})$
        \State \hspace{\algorithmicindent} $+ \beta \cdot \mathcal{L}_{\text{sim}}(h(\mathbf{x}^0_{\mathcal{B}}), h(\mathbf{x}_{\text{adv}}))$
        \State $\boldsymbol{\theta}_{\text{LoRA}} \leftarrow \boldsymbol{\theta}_{\text{LoRA}} - \eta_t \nabla_{\boldsymbol{\theta}_{\text{LoRA}}} \mathcal{L}_{\text{batch}}$ 
    \EndFor
\EndFor
\end{algorithmic}
\end{algorithm}

%% file: sec/results.tex
\begin{table*}[!ht]
\centering
\caption{Comparison of Clean and Adversarial Accuracy (\%). Our DAC-LoRA is compared against a standard CLIP-LoRA baseline and a naive PGD-LoRA adversarial baseline. The CLIP-LoRA baseline results are for a 4-shot setup. The highest adversarial accuracy for each setting is in \textbf{bold}. Values in \textcolor{red}{red} indicate training instability where model performance collapsed. A more extensive ablation study is also provided in the Supplementary Material (see table \ref{tab:ablation_results_final})}
\label{tab:final_results}
\begin{tabular}{@{}ll cc cc cc@{}}
\toprule
& & \multicolumn{2}{c}{\textbf{CLIP-LoRA} (Baseline)} & \multicolumn{2}{c}{\textbf{PGD-LoRA} (Baseline)} & \multicolumn{2}{c}{\textbf{DAC-LoRA} (Ours)} \\
\cmidrule(lr){3-4} \cmidrule(lr){5-6} \cmidrule(lr){7-8}
\textbf{Backbone} & \textbf{Dataset} & Clean & Adv & Clean & Adv & Clean & Adv \\
\midrule
\multirow{4}{*}{\textbf{ViT-B/16}} 
& Caltech101 & 95.16 & 29.19 & 82.60 & 61.01 & 94.20 & \textbf{72.86} \\
& DTD & 63.73 & 6.68 & 37.59 & 19.50 & 55.20 & \textbf{31.44} \\
& Oxford Pets & 80.99 & 10.08 & \textcolor{red}{4.80} & \textcolor{red}{3.24} & 81.98 & \textbf{44.45} \\
& UCF101 & 80.44 & 4.00 & \textcolor{red}{1.14} & \textcolor{red}{1.14} & 71.85 & \textbf{37.75} \\
\midrule
\multirow{4}{*}{\textbf{ViT-B/32}} 
& Caltech101 & 93.58 & 41.19 & 82.64 & 58.99 & 91.36 & \textbf{71.03} \\
& DTD & 60.10 & 10.10 & 36.82 & 17.79 & 47.99 & \textbf{29.37} \\
& Oxford Pets & 85.36 & 11.78 & \textcolor{red}{3.05} & \textcolor{red}{2.92} & 78.33 & \textbf{36.77} \\
& UCF101 & 74.55 & 7.89 & \textcolor{red}{0.93} & \textcolor{red}{0.69} & 65.08 & \textbf{33.33} \\
\bottomrule
\end{tabular}
\end{table*}

\section{Results}
\label{sec:results}

We evaluate our proposed framework, \textbf{DAC-LoRA}, on four diverse image classification datasets: Caltech101~\cite{FeiFei2004LearningGVCaltech101}, DTD~\cite{Cimpoi2013DescribingTIDTD}, Oxford Pets~\cite{Parkhi2012CatsADOxfordPets}, and UCF101~\cite{Soomro2012UCF101AD}. We compare its performance against two relevant baselines: a standard \textbf{CLIP-LoRA} fine-tuning on clean data and a naive \textbf{PGD-LoRA} adversarial fine-tuning. All models are evaluated for accuracy on the clean test set and robustness against a standard PGD attack, with specific parameters detailed in the Supplementary Material. 

The goal of this evaluation is not to establish a new state-of-the-art in adversarial defense, but to provide a clear proof-of-concept. We aim to demonstrate that the DAC framework can be easily integrated into a standard PEFT pipeline to significantly enhance robustness. The results in Table~\ref{tab:final_results} show that DAC-LoRA successfully achieves this, providing strong adversarial protection without significantly compromising accuracy.

The results presented in Table~\ref{tab:final_results} offer a clear narrative. While standard CLIP-LoRA achieves high clean accuracy, its robustness against adversarial attacks is consistently low, with adversarial accuracy often falling below 30\%. This establishes a baseline of a capable but fragile model.

Our experiments highlight a critical finding: naive adversarial fine-tuning with a method like PGD-LoRA can lead to profound instability. This is most evident on the Oxford Pets and UCF101 datasets, where the unstable results are marked in red. For the ViT-B/16 backbone, the PGD-LoRA baseline causes the clean accuracy on Oxford Pets to collapse from a high of 80.99\% (in the CLIP-LoRA baseline) to an unusable \textbf{4.80\%}. A similar catastrophic failure occurs on UCF101, where clean accuracy plummets from 80.44\% to a mere \textbf{1.14\%}. This demonstrates that without a careful approach, adversarial fine-tuning can catastrophically degrade a model's fundamental performance.

In stark contrast, our \textbf{DAC-LoRA} method successfully navigates this trade-off. In the same unstable scenarios, DAC-LoRA not only provides a massive boost in adversarial robustness but also preserves high clean accuracy. For Oxford Pets (ViT-B/16), it achieves \textbf{44.45\%} adversarial accuracy while maintaining a clean accuracy of 81.98\%. For UCF101, it improves adversarial accuracy from 4.00\% to \textbf{37.75\%} while keeping clean accuracy high at 71.85\%.

Even on datasets like Caltech101 and DTD where the PGD-LoRA baseline is more stable, our DAC-LoRA method provides a superior balance. On Caltech101 (ViT-B/16), it pushes adversarial accuracy to \textbf{72.86\%}, significantly outperforming both the standard CLIP-LoRA (29.19\%) and the naive PGD-LoRA (61.01\%), while maintaining a clean accuracy nearly identical to the best-performing baseline. These results collectively demonstrate that our dynamic curriculum is not just a method for improving robustness, but a crucial component for ensuring the stability and reliability of the fine-tuning process itself.

\section{Generalized DAC and Conclusion}
\label{sec:conclusion}

The core contribution of this work is a lightweight generalizable framework that we call the \textbf{Dynamic Adversarial Curriculum (DAC)}. The central idea, detailed in Algorithm~\ref{alg:generalized_dac}, is to decouple adversarial training from a fixed-strength attack. Instead, DAC provides a meta-algorithm that wraps around any iterative attack, using the FOSC score to dynamically increase the attack's potency. This ensures that the model first learns from weaker adversarial examples before being exposed to stronger ones, promoting stable and effective training.

This general framework, detailed in Algorithm\ref{alg:generalized_dac} (see Supplementary Material), decouples adversarial training from a fixed-strength attack. Our specific implementation, DAC-LoRA, serves as a successful proof-of-concept for this broader framework. The experimental results demonstrate that applying this curriculum to the parameter-efficient fine-tuning of CLIP models significantly boosts adversarial robustness without the catastrophic drop in clean-data accuracy often associated with naive adversarial training. This work highlights a promising and lightweight direction for building more reliable and secure fine-tuned models. We believe that our framework is a flexible tool that can be adapted to other attack modalities and architectures in future work, further enhancing the safety of AI systems.

%% file: sec/X_suppl.tex
\clearpage
\setcounter{page}{1}

\maketitlesupplementary

\section{Details on the FOSC Criterion}
\label{sec:supp_fosc}

Our method, DAC-LoRA, is built upon the principle of gradually increasing the difficulty of adversarial attacks during fine-tuning. This curriculum is governed by the \textbf{First-Order Stationary Condition (FOSC)}, a principled criterion for measuring the ``potency'' or convergence quality of an adversarial example~\cite{wang2019fosc}. This section provides a more detailed, intuitive explanation of this criterion.

The goal of adversarial training is to solve a min-max optimization problem~\cite{wang2019fosc}. In the context of our work, this can be formulated as finding the optimal LoRA parameters, $\theta_{\text{LoRA}}$, that minimize the training loss against the ``worst-case'' adversarial examples within a specified threat model:
\begin{equation}
\min_{\theta_{\text{LoRA}}} \mathbb{E}_{(x^0, y)\sim\mathcal{D}} \left[ \max_{x_{\text{adv}} \in \mathcal{B}_{\epsilon}(x^0)} \mathcal{L}(h(x_{\text{adv}}, \theta_{\text{LoRA}}), y) \right]
\label{eq:minmax}
\end{equation}
Here, the inner `max` operation aims to find the most effective adversarial example, $x_{\text{adv}}$, within an $\ell_\infty$-norm ball $\mathcal{B}_{\epsilon}$ of radius $\epsilon$ around the clean image $x^0$.

The FOSC criterion provides a quantitative measure of how well this inner maximization problem has been solved for a given adversarial example $x_{\text{adv}}$~\cite{wang2019fosc}. Intuitively, it measures the potential to further increase the loss from the current point $x_{\text{adv}}$ without violating the constraint. A smaller FOSC value indicates a stronger, better-converged attack, as there is less room for improvement.

The FOSC is formally defined as:
\begin{equation}
c(x_{\text{adv}}) = \max_{x \in \mathcal{B}_{\epsilon}(x^0)} \langle x - x_{\text{adv}}, \nabla_{x} \mathcal{L}(x_{\text{adv}}) \rangle
\end{equation}
where $\nabla_{x} \mathcal{L}(x_{\text{adv}})$ is the gradient of the loss with respect to the input image $x_{\text{adv}}$.

This expression has a convenient closed-form solution, which can be derived as follows~\cite{wang2019fosc}. By splitting the term $x - x_{\text{adv}}$ into $(x - x^0) + (x^0 - x_{\text{adv}})$, we can separate the maximization:
$$
c(x_{\text{adv}}) = \max_{x \in \mathcal{B}_{\epsilon}(x^0)} \langle x - x^0, \nabla_{x} \mathcal{L} \rangle + \langle x^0 - x_{\text{adv}}, \nabla_{x} \mathcal{L} \rangle
$$
The first term, $\max \langle x - x^0, \nabla_{x} \mathcal{L} \rangle$, is maximized when the perturbation $x - x^0$ is maximally aligned with the gradient. For the $\ell_\infty$-norm ball, this maximum value is equal to $\epsilon \|\nabla_{x} \mathcal{L}\|_1$. Combining the terms gives the final closed-form solution:
\begin{equation}
c(x_{\text{adv}}) = \epsilon \|\nabla_{x} \mathcal{L}(x_{\text{adv}})\|_1 - \langle x_{\text{adv}} - x^0, \nabla_{x} \mathcal{L}(x_{\text{adv}}) \rangle
\label{eq:fosc_closed}
\end{equation}
By using FOSC, we can create a principled \textbf{curriculum}~\cite{curr_learn}. Our dynamic training starts with a high FOSC threshold, meaning we only require ``easy'' or less-converged attacks. As training progresses, we linearly decrease this threshold, demanding progressively ``harder,'' better-converged adversarial examples (those with low FOSC scores)~\cite{wang2019fosc}. This ``easy-to-hard'' strategy stabilizes training and has been shown to improve final model robustness.

\section{A Generalized Dynamic Adversarial Curriculum}
\label{sec:generalized_dac}
The core contribution of our work is not a single implementation, but a generalizable framework we call the \textbf{Dynamic Adversarial Curriculum (DAC)}. The central idea is to decouple adversarial training from a fixed-strength attack. Instead, DAC provides a meta-algorithm that can wrap around any iterative attack method to progressively increase its strength based on a principled convergence criterion.

The DAC framework is detailed in Algorithm~\ref{alg:generalized_dac}. It operates on a generic principle of ``potency,'' represented by $\rho$. This potency is a hyperparameter of the underlying \texttt{AttackMethod} that controls its strength (e.g., the number of iterations for PGD, or the search steps for other attacks).

The procedure begins with the lowest potency and iteratively increases it as shown for PGD attack in \ref{fig:supp_dynamics}. At each step, it generates an adversarial example and evaluates its quality using the FOSC score. If the score is below the curriculum's current threshold, $c_t$, the attack is deemed ``potent enough'' for the current stage of training, and the generation process halts. If not, the potency is increased, and the attack is run again. This ensures that the model is never exposed to attacks that are unnecessarily strong for its current level of robustness, leading to more stable and effective training. This generalized approach allows practitioners to apply a curriculum to various types of iterative attacks beyond PGD, making it a flexible tool for improving adversarial robustness.

\begin{algorithm}[H]
\caption{Generalized FOSC-based Attack Curriculum}
\label{alg:generalized_dac}
\begin{algorithmic}[1]
\State \textbf{Input:} Model $h$, clean image $\mathbf{x}^0$, label $y$, FOSC threshold $c_t$.
\State \textbf{Hyperparameters:} An iterative \texttt{AttackMethod}, initial potency $\rho_0$, max potency $\rho_{\max}$, potency step size $\Delta\rho$.
\Procedure{Generate-Adversarial}{$\mathbf{x}^0, y, c_t$}
    \State $\rho \leftarrow \rho_0$ \Comment{Start with the lowest potency.}
    \State $\mathbf{x}_{\text{adv}} \leftarrow \mathbf{x}^0$
    \While{$\rho \le \rho_{\max}$}
        \State \Comment{Apply the attack with the current potency.}
        \State $\mathbf{x}_{\text{adv}} \leftarrow \texttt{AttackMethod}(\mathbf{x}^0, y, \rho)$
        \State Calculate FOSC score $c(\mathbf{x}_{\text{adv}})$ using Eq.~\eqref{eq:fosc}.
        \If{$c(\mathbf{x}_{\text{adv}}) < c_t$}
            \State \textbf{break} 
        \EndIf
        \State $\rho \leftarrow \rho + \Delta\rho$
    \EndWhile
    \State \textbf{return} $\mathbf{x}_{\text{adv}}$
\EndProcedure
\end{algorithmic}
\end{algorithm}

\section{Experimental Setup}
\label{sec:supp_experimental_setup}

All of our models were fine-tuned and evaluated using PyTorch on NVIDIA A6000 GPUs. The core DAC-LoRA method utilized a TRADES-inspired loss ($\beta=1.0$), a perturbation budget of $\epsilon=8/255$, and an initial FOSC threshold ($c_{\max}$) of 0.1. Finetuning was performed for 500 iterations with a batch size of 128 across all experiments.

\subsection{Setup for Main Comparison (Table 1)}
The results in Table 1 provide a direct comparison between our method and relevant baselines, all conducted in a \textbf{4-shot setting}. The CLIP-LoRA baseline was finetuned on clean data only, while the PGD-LoRA baseline was finetuned with a naive, fixed-strength PGD attack. Our DAC-LoRA model was trained using the dynamic curriculum with a training perturbation budget of $\epsilon=2/255$. For this comparison, all models were evaluated for robustness against a single, consistent PGD attack with \textbf{20 iterations} and a perturbation budget of $\epsilon=8/255$.

\subsection{Setup for Ablation Study (Table 2)}
To provide a more extensive analysis of our method, we conducted the ablation study presented in Table 2. This study investigates the behavior of only our DAC-LoRA method under different conditions. We varied two key hyperparameters: the number of few-shot examples (\textbf{4-shot vs. 16-shot}) and the training perturbation budget $\epsilon$ (\textbf{2/255 vs. 8/255}).

\begin{figure*}[t]
    \centering
    \includegraphics[width=\linewidth]{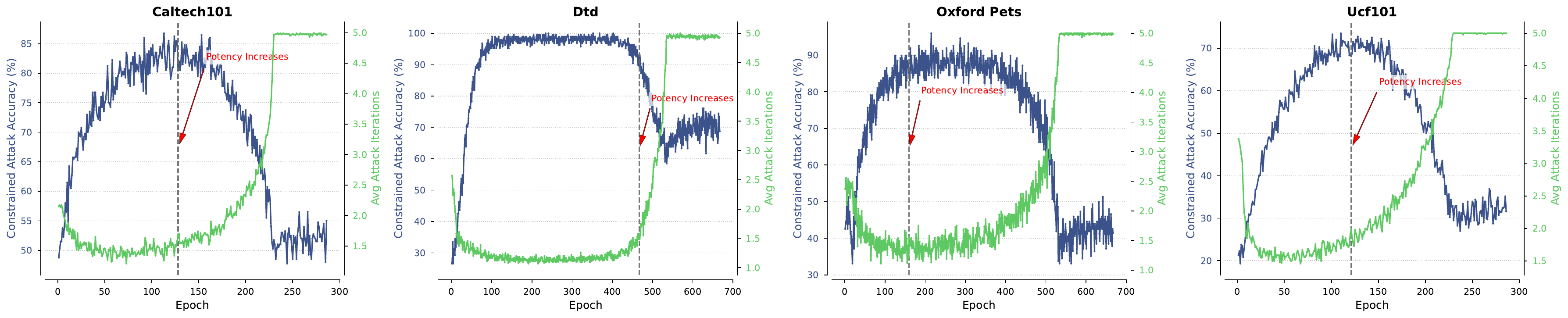}
    \includegraphics[width=\linewidth]{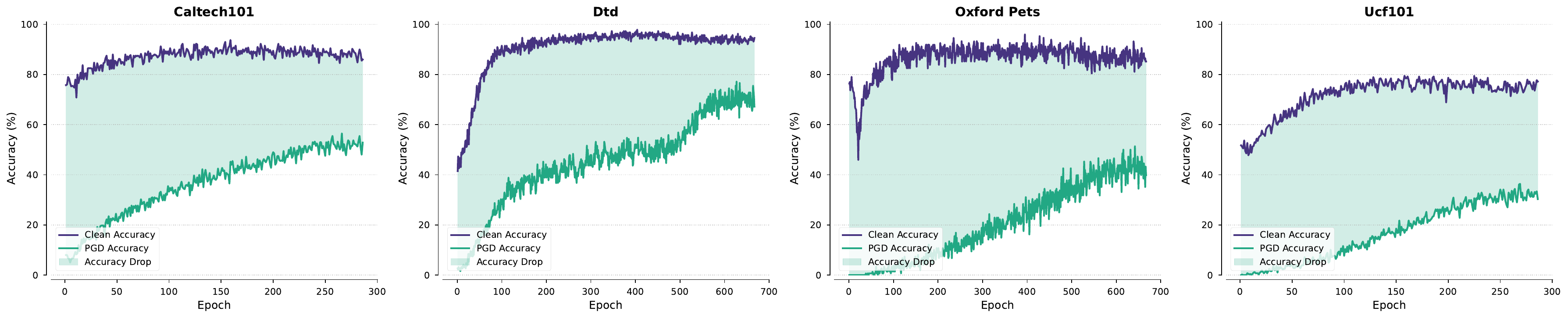}
    \caption{Training dynamics of DAC-LoRA with a ViT-B/16 backbone.
    \textbf{(Top Row):} The dynamic curriculum in action. As training progresses, the required attack potency increases, measured by the average PGD iterations (green line), resulting in improved accuracy against the curriculum's own attacks (blue line).
    \textbf{(Bottom Row):} Final model performance over epochs. Accuracy on clean data (blue line) remains high, while robustness against a fixed, strong PGD attack (green line) steadily increases.}
    \label{fig:supp_dynamics}
\end{figure*}
\subsection{Finetuning dynamics}
To provide a deeper understanding of the behavior of our DAC-LoRA framework, we visualize the key training dynamics for the ViT-B/16 backbone in Figure~\ref{fig:supp_dynamics}.

The plots in Figure~\ref{fig:supp_dynamics} illustrate the core principles of our method. The top row visualizes the dynamic curriculum itself. At the beginning of training, the FOSC threshold is high, so adversarial examples are generated with very few PGD steps (the green line is low). As training progresses and the model becomes more robust, the FOSC threshold decreases, forcing the attack to become more potent. This is reflected in the steady increase in the average number of attack iterations required to meet the FOSC criterion.

The bottom row shows the outcome of this curriculum. We plot the model's accuracy on clean data against its robustness to a fixed, strong PGD attack over the course of the training epochs. Across all four datasets, the clean accuracy (blue line) remains high and stable. Simultaneously, the adversarial accuracy (red line) shows a consistent and significant improvement. This visualization confirms that our DAC-LoRA framework successfully learns robust features without significantly compromising accuracy.

\begin{table*}[!ht]
\centering
\caption{Ablation Study of DAC-LoRA Finetuned Accuracy (\%). The table shows the performance of our method across 4-shot and 16-shot settings. For each setting, we report a single Clean Accuracy and the corresponding Adversarial Accuracy when trained and evaluated under two different perturbation budgets ($\epsilon=2/255$ and $\epsilon=8/255$).}
\label{tab:ablation_results_final}
\begin{tabular}{@{}ll c cc c cc@{}}
\toprule
& & \multicolumn{3}{c}{\textbf{4-Shot DAC-LoRA}} & \multicolumn{3}{c}{\textbf{16-Shot DAC-LoRA}} \\
\cmidrule(lr){3-5} \cmidrule(lr){6-8}
& & & \multicolumn{2}{c}{Adversarial Accuracy} & & \multicolumn{2}{c}{Adversarial Accuracy} \\
\cmidrule(lr){4-5} \cmidrule(lr){7-8}
\textbf{Backbone} & \textbf{Dataset} & \textbf{Clean Acc.} & \textbf{$\epsilon=2/255$} & \textbf{$\epsilon=8/255$} & \textbf{Clean Acc.} & \textbf{$\epsilon=2/255$} & \textbf{$\epsilon=8/255$} \\
\midrule
\multirow{4}{*}{\textbf{ViT-B/16}} 
& Caltech101 & 94.20 & 72.86 & 65.27 & 95.13 & 80.00 & 60.77 \\
& DTD & 55.20 & 31.44 & 22.93 & 63.36 & 41.78 & 22.46 \\
& Oxford Pets & 81.98 & 44.45 & 27.45 & 86.29 & 53.45 & 21.50 \\
& UCF101 & 71.85 & 37.75 & 26.38 & 75.10 & 47.29 & 26.20 \\
\midrule
\multirow{4}{*}{\textbf{ViT-B/32}} 
& Caltech101 & 91.36 & 71.03 & 60.53 & 92.82 & 76.92 & 57.04 \\
& DTD & 47.99 & 29.37 & 20.51 & 60.70 & 39.60 & 20.69 \\
& Oxford Pets & 78.33 & 36.77 & 21.04 & 78.90 & 46.12 & 16.60 \\
& UCF101 & 65.08 & 33.33 & 22.20 & 69.47 & 41.90 & 21.54 \\
\bottomrule
\end{tabular}
\end{table*}